\documentclass[11pt,a4paper]{article}
\usepackage[hyperref]{emnlp2020}
\usepackage{times}
\usepackage{latexsym}
\usepackage{amsmath}
\usepackage{bm}
\usepackage{multirow}

\usepackage{microtype}

\aclfinalcopy 

\usepackage{tikz}
\usepackage{color}
\usepackage{tikz-dependency}
\usetikzlibrary{shapes, arrows, positioning, decorations.markings}
\definecolor {processblue}{cmyk}{0.96,0,0,0}
\tikzstyle{int}=[draw, fill=blue!20, minimum size=2em]
\tikzstyle{init} = [pin edge={to-,thin,black}]
\usetikzlibrary{shapes}
\usetikzlibrary{fit}
\usetikzlibrary{chains}
\usetikzlibrary{arrows}
\tikzset{
semi/.style={
  semicircle, ,top color =white , bottom color = processblue!20 ,
draw, processblue , text=blue,
  draw,
  minimum size=0.3cm
  }
}
\tikzstyle{plate} = [draw, rectangle, rounded corners, fit=#1]
\tikzstyle{wrap} = [inner sep=0pt, fit=#1]
\tikzstyle{caption} = [node distance=0] %
\tikzstyle{bottom plate caption} = [caption, node distance=0, inner sep=0pt,
below left=-5pt and 0pt of #1.south east] %
\tikzstyle{top plate caption} = [caption, node distance=0, inner sep=0pt,
below left=0pt and 0pt of #1.north east] %

\usepackage{tikz}
\usetikzlibrary{shapes,arrows,shadows}
\usepackage{amsmath,bm,times}

\title{Recurrent Interaction Network for Jointly Extracting Entities and Classifying Relations}
\date{}
\author{
Kai Sun \\
 BDBC and SKLSDE\\
 Beihang University,China\\
  {\tt \small sunkai@buaa.edu.cn} \\
  \And
Richong Zhang\thanks{$\ \ $Corresponding author} \\
  BDBC and SKLSDE\\
  Beihang University, China\\
  {\tt\small zhangrc@act.buaa.edu.cn} \\
  \And
Samuel Mensah\\
  BDBC and SKLSDE\\
  Beihang University, China\\
    {\tt\small samensah@buaa.edu.cn} \\ 
    \AND
Yongyi Mao \\
  School of EECS\\
  University of Ottawa,Canada\\
  {\tt\small ymao@uottawa.ca} \\
      \And
Xudong Liu\\
  BDBC and SKLSDE\\
  Beihang University, Beijing, China\\
    {\tt\small liuxd@act.buaa.edu.cn} \\ \
  }
\begin{document}
\maketitle

\begin{abstract}

The idea of using multi-task learning approaches to address the joint extraction of entity and relation is motivated by the relatedness between the entity recognition task and the relation classification task. Existing methods using multi-task learning techniques to address the problem learn interactions among the two tasks through a shared network, where the shared information is passed into the task-specific networks for prediction. However, such an approach hinders the model from learning explicit interactions between the two tasks to improve the performance on the individual tasks. As a solution, we design a multi-task learning model which we refer to as recurrent interaction network which allows the learning of interactions dynamically, to effectively model task-specific features for classification. Empirical studies on two real-world datasets confirm the superiority of the proposed model.
 
\end{abstract}

\section{Introduction}\label{introduction}
The extraction of entities and relations from textual data comprises of two sub-tasks: entity recognition (ER) and relation classification (RC). The ER task aims at extracting all entities in a given text. The RC task aims at classifying the relation between any pair of entities in the text. In practice, both tasks are required to be solved jointly, and have been observed to contribute significantly in extracting structured knowledge from unstructured text for several applications, including knowledge base construction~\cite{DBLP:conf/acl/KomninosM17,DBLP:conf/aaai/DengXLYDFLS19,DBLP:conf/acl/NathaniCSK19}. For instance, consider the sentence \emph{John was born in Sheffield, a city in the north of England}. The goal of a joint entity and relation extraction task is to identify all the factual relational triples (or relational facts) \emph{\tt (Sheffield, birth\_place\_of, John)} and \emph{\tt (England, contains, Sheffield)}.


\begin{figure}[htbp]
\centering
	\renewcommand\tabcolsep{10pt}
	\begin{tabular}{c c}
	\scalebox{0.7}{    \begin{tikzpicture}[-latex ,auto ,node distance =2 cm and 2 cm ,on grid ,
semithick , 
state/.style ={ circle ,top color =white , bottom color = processblue!20 ,
draw, processblue , text=black , minimum width =0.025cm, text width=0.025cm},
box/.style ={rectangle ,top color =white , bottom color = processblue!20 ,
draw, processblue , text=blue , minimum width =0.5cm , minimum height = 0.5cm},
dbox/.style ={rectangle ,top color =white , bottom color = processblue!20 ,
draw, processblue , text=blue , minimum width =0.5cm , minimum height = 0.5cm},
neuron/.style ={rectangle ,top color =white , bottom color = red!20 ,
draw, red , text=red , minimum width =0.5cm , minimum height = 0.5cm, rounded corners},
triangle/.style = {top color =white , bottom color = processblue!20 ,
draw, processblue , text=blue, regular polygon, regular polygon sides=3, minimum size=0.5cm, draw },
node rotated/.style = {rotate=270},
    border rotated/.style = {shape border rotate=270},
gn/.style={trapezium, trapezium angle=67.5, draw, inner ysep=5pt, outer sep=0pt, minimum height=1.81mm, minimum width=0pt}
    ]
    


   

\node [] (S1) at (-2, -2.75) {$S$};
    \node [box] (g11) at (-0.5, -2) {$A$};
    \node [box] (g12) at (-0.5, -3.5) {$B$};
   
   \draw [line width=0.8pt, rounded corners=0.5mm](S1)--(g11);
   \draw [line width=0.8pt, rounded corners=0.5mm](S1)--(g12);
   
    \end{tikzpicture}} &
	\scalebox{0.7}{    \begin{tikzpicture}[-latex ,auto ,node distance =2 cm and 2 cm ,on grid ,
semithick , 
state/.style ={ circle ,top color =white , bottom color = processblue!20 ,
draw, processblue , text=black , minimum width =0.025cm, text width=0.025cm},
box/.style ={rectangle ,top color =white , bottom color = processblue!20 ,
draw, processblue , text=blue , minimum width =0.5cm , minimum height = 0.5cm},
dbox/.style ={rectangle ,top color =white , bottom color = processblue!20 ,
draw, processblue , text=blue , minimum width =0.5cm , minimum height = 0.5cm},
neuron/.style ={rectangle ,top color =white , bottom color = red!20 ,
draw, red , text=red , minimum width =0.5cm , minimum height = 0.5cm, rounded corners},
triangle/.style = {top color =white , bottom color = processblue!20 ,
draw, processblue , text=blue, regular polygon, regular polygon sides=3, minimum size=0.5cm, draw },
node rotated/.style = {rotate=270},
    border rotated/.style = {shape border rotate=270},
gn/.style={trapezium, trapezium angle=67.5, draw, inner ysep=5pt, outer sep=0pt, minimum height=1.81mm, minimum width=0pt}
    ]

    \node [] (S1) at (-2, -2.75) {$S$};
    \node [box] (g11) at (-0.5, -2) {$A$};
    \node [box] (g12) at (-0.5, -3.5) {$B$};
   
   \draw [line width=0.8pt, rounded corners=0.5mm](S1)--(g11);
   \draw [line width=0.8pt, rounded corners=0.5mm](S1)--(g12);
    
    \draw [line width=0.8pt, rounded corners=0.5mm](-0.6,-2.25)--(-0.6,-3.25);
    
    \draw [line width=0.8pt, rounded corners=0.5mm](-0.4,-3.25)--(-0.4,-2.25);

        \end{tikzpicture}}
	\\
	  	(a) Flat Structure &
	  (b) Graph Structure
	\end{tabular}
	\caption{Two topological structures for multi-task learning. Here, $A$ and $B$ are related tasks, and $S$ is the shared information of the two tasks. The directed edges define the information flow.}
	\label{fig:motivation}
\end{figure}
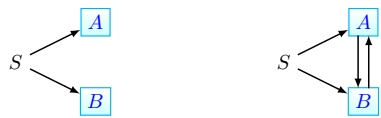

The simplest approach to solve this joint task is to utilize a pipeline-based approach by firstly extracting all entities in the sentence and then classifying the relation between all entity pairs~\cite{DBLP:journals/jmlr/ZelenkoAR03,DBLP:conf/acl/ZhouSZZ05,DBLP:conf/acl/ChanR11}. However, pipeline-based approaches disregard the correlation between ER and RC tasks, leading to error propagation in these methods.

Recently, researchers have exploited multi-task learning (MTL)~\cite{DBLP:conf/icml/CollobertW08} techniques to capture the correlation between the ER and RC tasks, and have successfully improved the performance of the individual tasks~\cite{DBLP:conf/acl/MiwaB16,DBLP:conf/acl/FuLM19,zeng2019copymtl}. These methods have a flat structure~\cite{DBLP:conf/aaai/LiuFDQC19}. Figure~\ref{fig:motivation}(a) shows a flat structure for multi-task learning. Methods using a flat structure learn interactions between tasks through a shared network, and extract a shared common representation which is exploited by task-specific networks independently. We refer to MTL methods utilizing a flat structure as conventional MTL methods. A conventional MTL method is effective to an extent because they help to improve generalization performance on all the tasks. However, it is based on the strong assumption that the shared network is sufficient to capture the correlations between the tasks. 





Even so, identifying the relational facts in sentences is a difficult problem. Reason being that several relational facts may overlap in a sentence~\cite{DBLP:conf/emnlp/Zhang0M18}. Although a conventional MTL method may learn task-specific features and has been successfully applied in a wide variety of scenarios~\cite{DBLP:conf/ijcai/ZhangW16a,DBLP:conf/emnlp/WuSCHS16,DBLP:conf/naacl/GooGHHCHC18,DBLP:conf/emnlp/HanNP19,DBLP:conf/aclnut/LiBZL19,DBLP:conf/emnlp/NishinoMKTMO19,DBLP:conf/acl/LiuHCG19,DBLP:conf/acl/HuPHLL19}, its flat structure restricts the model to effectively learn the correlations between tasks. For example in Figure~\ref{fig:motivation}(a), the model cannot explicitly learn correlations between the two tasks. Without modeling explicit interactions, as shown in a sequence learning task~\cite{DBLP:conf/aaai/LiuFDQC19}, the existing MTL-based methods~\cite{DBLP:conf/acl/MiwaB16,DBLP:conf/acl/FuLM19,zeng2019copymtl} cannot effectively capture the correlation between the ER and the RC tasks. 


In this paper, we overcome the aforementioned limitation of previous MTL-based methods by proposing a recurrent interaction network (RIN) to effectively capture the correlations between the ER and RC tasks. RIN has a multi-task learning architecture which allows interactions between the ER and RC tasks to be learned explicitly, with the aim to improve the performance on the individual tasks.  More specifically, RIN has a recurrent structure comprising of multiple interaction layers, allowing the model to progressively learn complex interactions while refining predictions for ER and RC. The RIN structure is an example of a multi-task learning network with a graph structure~\cite{DBLP:conf/aaai/LiuFDQC19}. We show the graph structure in Figure~\ref{fig:motivation}(b). As shown by our experiment, the proposed model progressively provides discriminating features which is an essential requirement for the individual task for classification. Empirical studies on NYT and WebNLG datasets achieve new state-of-the-art performances and confirm the effectiveness of the proposed RIN model.

\section{Related Work}\label{relworks}

Previous neural methods proposed for jointly extracting entities and relations can generally be categorized into three classes. The first class models the joint extraction task as a sequence labeling problem~\cite{DBLP:conf/acl/ZhengWBHZX17,DBLP:conf/aaai/DaiXLDSW19,DBLP:conf/aaai/TakanobuZLH19,DBLP:journals/corr/abs-1909-04273}. Among the proposed works, \cite{DBLP:conf/acl/ZhengWBHZX17} was the first to introduce a tagging strategy to address the problem, transferring the joint extraction task to a sequence labelling problem.
However, this method has the fundamental weakness of addressing the overlapping problem of relational facts in the text. To meet it, \cite{DBLP:conf/aaai/DaiXLDSW19} proposed a position-attentive tagging scheme to solve the overlapping problem. Meanwhile, \cite{DBLP:conf/aaai/TakanobuZLH19, DBLP:journals/corr/abs-1909-04273}
approach the problem by decomposing the joint extraction task into two sequence labeling sub-tasks, to address the joint entity and relation extraction problem.



The second class of works use a sequence-to-sequence (seq2seq) approach to address the problem~\cite{DBLP:conf/acl/LiuZZHZ18,zeng-etal-2019-learning}. \citep{DBLP:conf/acl/LiuZZHZ18} employs a seq2seq model to directly extract relational facts from the sentence by decoding the first entity, second entity, and relation in that order. But, their approach is limited to extracting a predefined number of relational facts from the text. In extracting relational triples, the order of extraction is key to identify the relational facts. As such, \cite{zeng-etal-2019-learning} proposed a seq2seq approach which utilizes a reinforcement learning model to learn the order of extracting the relational triples. Although effective, the proposed seq2seq models~\cite{DBLP:conf/acl/LiuZZHZ18,zeng-etal-2019-learning} only decode a single word for an entity.

The third class design a multi-task learning model to extract relational facts. Only few works using this approach have been proposed~\cite{DBLP:conf/acl/MiwaB16,DBLP:conf/acl/FuLM19,zeng2019copymtl}. \cite{DBLP:conf/acl/MiwaB16} is one of the first works to extract relational facts using an MTL framework. 
\citep{zeng2019copymtl} proposed an MTL model which comprises of an ER model to extract entities with multi-tokens, and a seq2seq model to extract relational facts. Their approach solves the entity extraction problem faced by models which are solely seq2seq based. \citep{DBLP:conf/acl/FuLM19} exploited a bidirectional recurrent neural network and graph convolutional network to extract common features of the ER and RC tasks, which are further fed into two independent classifiers for ER and RC predictions.  Despite the substantial efforts and great successes in the design of these MTL-based methods, these methods follow the conventional MTL approach~\cite{DBLP:conf/icml/CollobertW08}. Thus, they only capture implicit interactions by means of the shared network of the ER and RC tasks. 



Modelling explicit interactions between multiple tasks in an MTL architecture has been explored to improve predictions in several domains ~\cite{DBLP:conf/acl/HeLND19,DBLP:conf/aaai/Zhao0ZW19,DBLP:conf/emnlp/DankersRLS19,DBLP:conf/emnlp/LanWWNW17,DBLP:conf/aaai/LiuFDQC19,DBLP:conf/ijcai/LiuQH16}.
As mentioned in Section~\ref{introduction}, it is difficult to effectively learn the correlations between the ER and the RC tasks. To this end, we follow some of the ideas from other domains to dynamically learn the interactions between the two tasks, refining the classifiers of the tasks. To the best of our knowledge, this is the first work to model explicit interactions in a multi-task learning architecture for the joint extraction of entities and relations in text. 
\begin{figure*}[htbp]
	\centering
	\renewcommand\tabcolsep{1.5pt}
	\begin{tabular}{c c c}
	\scalebox{0.6}{\begin{tikzpicture}[-latex ,auto ,node distance =2 cm and 2 cm ,on grid ,
semithick , 
state/.style ={ circle ,top color =white , bottom color = processblue!20 ,
draw, processblue , text=black , minimum width =0.025cm, text width=0.025cm},
box/.style ={rectangle ,top color =white , bottom color = processblue!20 ,
draw, processblue , text=blue , minimum width =0.5cm , minimum height = 0.5cm},
dbox/.style ={rectangle ,top color =white , bottom color = processblue!20 ,
draw, processblue , text=blue , minimum width =0.5cm , minimum height = 0.5cm},
neuron/.style ={rectangle ,top color =white , bottom color = red!20 ,
draw, red , text=red , minimum width =0.5cm , minimum height = 0.5cm, rounded corners},
triangle/.style = {top color =white , bottom color = processblue!20 ,
draw, processblue , text=blue, regular polygon, regular polygon sides=3, minimum size=0.5cm, draw },
node rotated/.style = {rotate=270},
    border rotated/.style = {shape border rotate=270},
gn/.style={trapezium, trapezium angle=67.5, draw, inner ysep=5pt, outer sep=0pt, minimum height=1.81mm, minimum width=0pt}
    ]

    \node [] (H0) at (0, 0){$H^{(0)}$};
    \node [box](Cr0) at (0.7, 1.){$C_{r}$};
    \node [box](Cg0) at (0.7, -1.){$C_{e}$};
    
    \node [style=neuron] (GRUr0) at (2.25, 1) {$\rm GRU_{r}$};
    \node [style=neuron] (GRUg0) at (2.25, -1) {$\rm GRU_{e}$};
    
    \node [box](Cr1) at (3.9, 1){$C_{r}$};
    \node [box](Cg1) at (3.9, -1){$C_{e}$};
    
    \node [style=neuron] (GRUr1) at (5.35, 1) {$\rm GRU_{r}$};
    \node [style=neuron] (GRUg1) at (5.35, -1) {$\rm GRU_{e}$};
    
    \node [] (d1) at (6.75, 1){\textbf{...}};
    \node [] (d2) at (6.75, -1){\textbf{...}};
    \node [] (d3) at (6.75, 0){\textbf{...}};
    
    \node [style=neuron] (GRUrk) at (8.1, 1) {$\rm GRU_{r}$};
    \node [style=neuron] (GRUgk) at (8.1, -1) {$\rm GRU_{e}$};

    \node [box](Crk) at (9.6, 1){$C_{r}$};
    \node [box](Cgk) at (9.6, -1){$C_{e}$};
    
    \node [] (H1) at (2.25, 0){$H^{(0)}$};
    \node [] (H2) at (5.35, 0){$H^{(1)}$};
    \node [] (Hk) at (8.1, 0){$H^{(k-1)}$};
    
    \node [] (Yr0) at (1.4, 1.35){$Y_r^{(0)}$};
    \node [] (Yr1) at (4.6, 1.35){$Y_r^{(1)}$};
    \node [] (Yrk_1) at (7.5, 1.35){$Y_r^{(k-1)}$};
    \node [] (Yrk) at (10.4, 1.35){$Y_r^{(k)}$};
    
    \node [] (Yg0) at (1.4, -0.65){$Y_e^{(0)}$};
    \node [] (Yg1) at (4.6, -0.65){$Y_e^{(1)}$};
    \node [] (Ygk_1) at (7.5, -0.65){$Y_e^{(k-1)}$};
    \node [] (Ygk) at (10.4, -0.65){$Y_e^{(k)}$};
    
    \node [state](o0) at (3.1,0){};
    \node []() at (3.1, 0){$+$};
    
    \node [state](o1) at (6.2,0){};
    \node []() at (6.2, 0){$+$};
    
    \node [] (Hr1) at (3.25, 1.35){$H_r^{(1)}$};
    \node [] (Hr2) at (6.3, 1.35){$H_r^{(2)}$};
    \node [] (Hrk) at (9.05, 1.35){$H_r^{(k)}$};
    
    \node [] (He1) at (3.1, -1.35){$H_e^{(1)}$};
    \node [] (He2) at (6.2, -1.35){$H_e^{(2)}$};
    \node [] (Hek) at (8.92, -1.35){$H_e^{(k)}$};
    
    \draw [line width=0.8pt, rounded corners=0.5mm](H0)--(0,1)--(Cr0);
    \draw [line width=0.8pt, rounded corners=0.5mm](H0)--(0,-1)--(Cg0);
    
    \draw [line width=0.8pt, rounded corners=0.5mm](Cr0)--(GRUr0);
    \draw [line width=0.8pt, rounded corners=0.5mm](Cg0)--(GRUg0);
    
    \draw [line width=0.8pt, rounded corners=0.5mm](GRUr0)--(Cr1);
    \draw [line width=0.8pt, rounded corners=0.5mm](GRUg0)--(Cg1);
    
    \draw [line width=0.8pt, rounded corners=0.5mm](GRUr1)--(6.6,1);
    \draw [line width=0.8pt, rounded corners=0.5mm](GRUg1)--(6.6,-1);

    \draw [line width=0.8pt, rounded corners=0.5mm](Cr1)--(GRUr1);
    \draw [line width=0.8pt, rounded corners=0.5mm](Cg1)--(GRUg1);

    \draw [line width=0.8pt, rounded corners=0.5mm](H1)--(GRUr0);
    \draw [line width=0.8pt, rounded corners=0.5mm](H1)--(GRUg0);
    
    \draw [line width=0.8pt, rounded corners=0.5mm](H2)--(GRUr1);
    \draw [line width=0.8pt, rounded corners=0.5mm](H2)--(GRUg1);

    \draw [line width=0.8pt, rounded corners=0.5mm](6.9,1)--(GRUrk);
    \draw [line width=0.8pt, rounded corners=0.5mm](6.9,-1)--(GRUgk);

    \draw [line width=0.8pt, rounded corners=0.5mm](H0)--(H1);
    \draw [line width=0.8pt, rounded corners=0.5mm](2.6,0)--(o0);
    \draw [line width=0.8pt, rounded corners=0.5mm](o0)--(H2);
    \draw [line width=0.8pt, rounded corners=0.5mm](5.7,0)--(o1);
    \draw [line width=0.8pt, rounded corners=0.5mm](o1)--(6.63,0);
    \draw [line width=0.8pt, rounded corners=0.5mm](6.9,0)--(Hk);
    
    \draw [line width=0.8pt, rounded corners=0.5mm](Hk)--(GRUrk);
    \draw [line width=0.8pt, rounded corners=0.5mm](Hk)--(GRUgk);
    
    \draw [line width=0.8pt, rounded corners=0.5mm](GRUrk)--(Crk);
     \draw [line width=0.8pt, rounded corners=0.5mm](GRUgk)--(Cgk);
     
     \draw [line width=0.8pt, rounded corners=0.5mm](Crk)--(10.7,1);
     \draw [line width=0.8pt, rounded corners=0.5mm](Cgk)--(10.7,-1);
     
     \draw [line width=0.8pt, rounded corners=0.5mm](3.1,1)--(o0);
     \draw [line width=0.8pt, rounded corners=0.5mm](3.1,-1)--(o0);
     
     \draw [line width=0.8pt, rounded corners=0.5mm](6.2,1)--(o1);
     \draw [line width=0.8pt, rounded corners=0.5mm](6.2,-1)--(o1);
    
    \end{tikzpicture}}
	&
	\scalebox{0.6}{\begin{tikzpicture}[-latex ,auto ,node distance =2 cm and 2 cm ,on grid ,
semithick , 
state/.style ={ circle ,top color =white , bottom color = processblue!20 ,
draw, processblue , text=black , minimum width =0.025cm, text width=0.025cm},
box/.style ={rectangle ,top color =white , bottom color = processblue!20 ,
draw, processblue , text=blue , minimum width =0.5cm , minimum height = 0.5cm},
dbox/.style ={rectangle ,top color =white , bottom color = processblue!20 ,
draw, processblue , text=blue , minimum width =0.5cm , minimum height = 0.5cm},
neuron/.style ={rectangle ,top color =white , bottom color = red!20 ,
draw, red , text=red , minimum width =0.5cm , minimum height = 0.5cm, rounded corners},
triangle/.style = {top color =white , bottom color = processblue!20 ,
draw, processblue , text=blue, regular polygon, regular polygon sides=3, minimum size=0.5cm, draw },
node rotated/.style = {rotate=270},
    border rotated/.style = {shape border rotate=270},
gn/.style={trapezium, trapezium angle=67.5, draw, inner ysep=5pt, outer sep=0pt, minimum height=1.81mm, minimum width=0pt}
    ]
    \node [] (h1) at (-0.25, 0){$h_1$};
    \node [] (h2) at (-0.25, -1){$h_2$};
    \node [] (h3) at (-0.25, -2){$h_3$};
    \node [] (wg) at (-0.5, 0.5){$W_e$};
    \node [] (yg1) at (3.55, 0){$y_1$};
    \node [] (yg2) at (3.55, -1){$y_2$};
    \node [] (yg3) at (3.55, -2){$y_3$};

    \node [state](m1) at (1,0){};
    \node []() at (1,0){$*$};
    \node [state](m2) at (1,-1){};
    \node []() at (1,-1){$*$};
    \node [state](m3) at (1,-2){};
    \node []() at (1,-2){$*$};
    
    \node [box, rotate=-90, minimum width = 2.5cm] (att) at (2.15, -1) {softmax};
    \draw [line width=0.8pt, rounded corners=1mm](h1)--(m1);
    \draw [line width=0.8pt, rounded corners=1mm](h2)--(m2);
    \draw [line width=0.8pt, rounded corners=1mm](h3)--(m3);
    \draw [line width=0.8pt, rounded corners=1mm](m1)--(1.9,0);
    \draw [line width=0.8pt, rounded corners=1mm](m2)--(1.9,-1);
    \draw [line width=0.8pt, rounded corners=1mm](m3)--(1.9,-2);
    \draw [line width=0.8pt, rounded corners=1mm](2.4,0)--(3.2,0);
    \draw [line width=0.8pt, rounded corners=1mm](2.4,-1)--(3.2,-1);
    \draw [line width=0.8pt, rounded corners=1mm](2.4,-2)--(3.2,-2);
    
    \draw [line width=0.8pt, rounded corners=0.2mm](wg)--(1,0.5)--(m1.north);
    \draw [line width=0.8pt, rounded corners=0.2mm](wg.south)--(-0.5,-0.5)--(1,-0.5)--(m2.north);
    \draw [line width=0.8pt, rounded corners=0.2mm](wg.south)--(-0.5,-1.5)--(1,-1.5)--(m3.north);
    \end{tikzpicture}}&\scalebox{0.6}{\begin{tikzpicture}[-latex ,auto ,node distance =2 cm and 2 cm ,on grid ,
semithick , 
state/.style ={ circle ,top color =white , bottom color = processblue!20 ,
draw, processblue , text=black , minimum width =0.025cm, text width=0.025cm},
box/.style ={rectangle ,top color =white , bottom color = processblue!20 ,
draw, processblue , text=blue , minimum width =0.5cm , minimum height = 0.5cm},
dbox/.style ={rectangle ,top color =white , bottom color = processblue!20 ,
draw, processblue , text=blue , minimum width =0.5cm , minimum height = 0.5cm},
neuron/.style ={rectangle ,top color =white , bottom color = red!20 ,
draw, red , text=red , minimum width =0.5cm , minimum height = 0.5cm, rounded corners},
triangle/.style = {top color =white , bottom color = processblue!20 ,
draw, processblue , text=blue, regular polygon, regular polygon sides=3, minimum size=0.5cm, draw },
node rotated/.style = {rotate=270},
    border rotated/.style = {shape border rotate=270},
gn/.style={trapezium, trapezium angle=67.5, draw, inner ysep=5pt, outer sep=0pt, minimum height=1.81mm, minimum width=0pt}
    ]
     \node [] (h1) at (-0.25, 0){$h_1$};
    \node [] (h2) at (-0.25, -1){$h_2$};
    \node [] (h3) at (-0.25, -2){$h_3$};
    \node [] (h12) at (-0.5, 0.5){$h_1$};
    
    \node [state](m1) at (1,0){};
    \node []() at (1,0){$\oplus$};
    \node [state](m2) at (1,-1){};
    \node []() at (1,-1){$\oplus$};
    \node [state](m3) at (1,-2){};
    \node []() at (1,-2){$\oplus$};
    
    \node [state](c1) at (2,0){};
    \node []() at (2,0){$*$};
    \node [state](c2) at (2,-1){};
    \node []() at (2,-1){$*$};
    \node [state](c3) at (2,-2){};
    \node []() at (2,-2){$*$};
    
    \node [state](c4) at (3,0){};
    \node []() at (3,0){$\phi$};
    \node [state](c5) at (3,-1){};
    \node []() at (3,-1){$\phi$};
    \node [state](c6) at (3,-2){};
    \node []() at (3,-2){$\phi$};
    
    \node [state](c7) at (4,0){};
    \node []() at (4,0){$*$};
    \node [state](c8) at (4,-1){};
    \node []() at (4,-1){$*$};
    \node [state](c9) at (4,-2){};
    \node []() at (4,-2){$*$};
    
    \node [state](c10) at (5,0){};
    \node []() at (5,0){$\sigma$};
    \node [state](c11) at (5,-1){};
    \node []() at (5,-1){$\sigma$};
    \node [state](c12) at (5,-2){};
    \node []() at (5,-2){$\sigma$};
    
    \node [] (wm1) at (2, 0.6){$W_m$};
    \node [] (wm2) at (2, -0.4){$W_m$};
    \node [] (wm3) at (2, -1.4){$W_m$};
    
    \node [] (wr1) at (4, 0.6){$W_r$};
    \node [] (wr2) at (4, -0.4){$W_r$};
    \node [] (wr3) at (4, -1.4){$W_r$};
    
    \node [] (yr1) at (6.3, 0.){$y_{(1,1)}$};
    \node [] (yr2) at (6.3, -1){$y_{(1,2)}$};
    \node [] (yr3) at (6.3, -2){$y_{(1,3)}$};

    \draw [line width=0.8pt, rounded corners=1mm](h1)--(m1);
    \draw [line width=0.8pt, rounded corners=1mm](h2)--(m2);
    \draw [line width=0.8pt, rounded corners=1mm](h3)--(m3);
    
    \draw [line width=0.8pt, rounded corners=0.2mm](h12)--(1,0.5)--(m1.north);
    \draw [line width=0.8pt, rounded corners=0.2mm](h12.south)--(-0.5,-0.5)--(1,-0.5)--(m2.north);
    \draw [line width=0.8pt, rounded corners=0.2mm](h12.south)--(-0.5,-1.5)--(1,-1.5)--(m3.north);
    
    \draw [line width=0.8pt, rounded corners=0.2mm](m1)--(c1);
    \draw [line width=0.8pt, rounded corners=0.2mm](m2)--(c2);
    \draw [line width=0.8pt, rounded corners=0.2mm](m3)--(c3);
    
    \draw [line width=0.8pt, rounded corners=1mm](2,0.5)--(c1);
    \draw [line width=0.8pt, rounded corners=1mm](2,-0.5)--(c2);
    \draw [line width=0.8pt, rounded corners=1mm](2,-1.5)--(c3);
    
    \draw [line width=0.8pt, rounded corners=1mm](4,0.5)--(c7);
    \draw [line width=0.8pt, rounded corners=1mm](4,-0.5)--(c8);
    \draw [line width=0.8pt, rounded corners=1mm](4,-1.5)--(c9);
    
    \draw [line width=0.8pt, rounded corners=1mm](c1)--(c4);
    \draw [line width=0.8pt, rounded corners=1mm](c2)--(c5);
    \draw [line width=0.8pt, rounded corners=1mm](c3)--(c6);
    
    \draw [line width=0.8pt, rounded corners=1mm](c4)--(c7);
    \draw [line width=0.8pt, rounded corners=1mm](c5)--(c8);
    \draw [line width=0.8pt, rounded corners=1mm](c6)--(c9);
    
    \draw [line width=0.8pt, rounded corners=1mm](c7)--(c10);
    \draw [line width=0.8pt, rounded corners=1mm](c8)--(c11);
    \draw [line width=0.8pt, rounded corners=1mm](c9)--(c12);
    
    \draw [line width=0.8pt, rounded corners=1mm](c10)--(5.8,0);
    \draw [line width=0.8pt, rounded corners=1mm](c11)--(5.8,-1);
    \draw [line width=0.8pt, rounded corners=1mm](c12)--(5.8,-2);
    \end{tikzpicture}}
	\\
	(a) Framework of RIN.& (b) The ER module ($C_e$). & (c)  The RC module ($C_r$).
	\end{tabular}
	\caption{(a) The framework of RIN. (b) The entitiy recognition module. (c) The relation classification module. In (b) and (c), we use a toy example of shared features $H=\{h_1, h_2, h_3\}$ to demonstrate the entity prediction for word $w_i$ and relation prediction for all pairs $(w_1,w_1),(w_1,w_2),(w_1,w_3)$. $+$, $\oplus$, $*$, $\phi$, and $\sigma$ denote a summation operator, a concatenation operator, a matrix multiplication, relu activation function and sigmoid function respectively.}
	\label{fig:our_model}
\end{figure*}
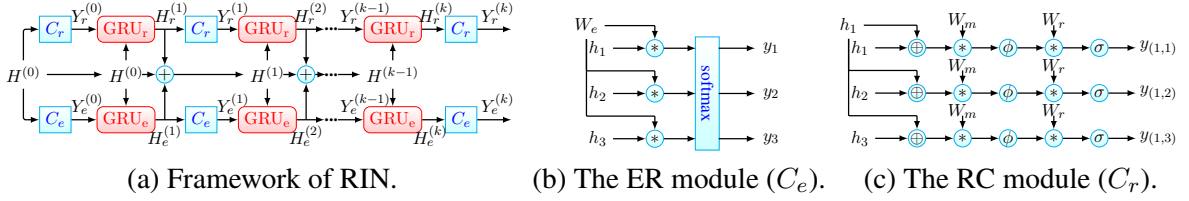

\section{Problem Statement}
In this section, we formally describe the joint entity and relation extraction problem. For a set $T=\{t_1,\cdots, t_l\}$ of pre-defined $l$ relation types, and a given sentence $s=\{w_1,w_2,\cdots,w_n\}$ of $n$ words, the problem is to extract all relational facts for the given sentence. In this paper, a single relational triple is of the form $\left<w_i,t,w_j\right>$, where $w_i,w_j\in s$ are entity words or heads of multi-token entities, and $w_i\neq w_j$, and the relation $t\in T$. The goal is to predict the probability $y_{(i,j)}^t$ that the relational triple $\left<w_i,t,w_j\right>$ is factual given the word pair $(w_i,w_j)$. Besides, the entity recognition task can identify the head and tail words of multi-token entities for the extracted relational triple.


\section{Model}\label{model}
In this section, we describe the recurrent interaction network (RIN) for extracting relational facts in text. The RIN model is composed of an entity recognition (ER) module and a relation classification (RC) module. We start by presenting an overview of the RIN model, showing the interaction between the ER and RC tasks. Next, we elaborate the ER and RC modules and define the training objective. The framework of RIN is shown in Figure~\ref{fig:our_model}.

\subsection{Recurrent Interaction Network}
The RIN model we propose uses a bidirectional LSTM network to learn correlations between the ER and the RC tasks, and derives shared features for the two tasks. We denote $H$ as the output of shared features, where $H=\{h_1, h_2, \ldots,h_n\}$ corresponds to the representations of words in sentence $s$. A straightforward strategy for the joint ER and RC task is to pass $H$ into independent ER or RC modules for predictions. Denote $C_e$ as the ER module to identify and extract entities in the text, and $C_r$ as the RC module to classify relational triples in the text. Formally, $Y_e$ and $Y_r$, the set of predictions of the entities and relational triples are formulated as:
\begin{equation}
\begin{aligned}
Y_r &= C_r(H) \\
Y_e &= C_e(H)     
\end{aligned}
\end{equation}
where $Y_e=: \{y_i | h_i \in H\}$, $Y_r=: \{y_{(i,j)} | h_i,h_j\in H \}$, $y_i$ is a probability distribution over BIOES labels~\cite{DBLP:conf/acl/FuLM19}, and $y_{(i,j)}$ is a probability distribution over the relation types $t\in T$.  This structure is basically a conventional MTL method, where interactions are learned implicitly, impeding dynamic learning of intrinsic correlations between the two tasks. 

To enhance the interaction between the two tasks, we dynamically learn the explicit interactions between the ER and RC tasks. Each layer of the RIN model is an interaction layer comprising of two separate gated recurrent units (GRUs), accounting for the ER task and the RC task. The GRU networks are designed to model task-specific features at the $k$-th layer, taking into account the previous shared features $H^{(k-1)}$ and the previous predictions $Y_e^{k-1}$ and $Y_r^{k-1}$. Meanwhile, the shared features $H^{(k)}$ generated at the $k$-th layer is a sum of the previous task-specific features and the previous shared features $H^{(k-1)}$. Such a mechanism ensures that we retain the learned correlations as learning progresses along the network.

Let \text{GRU}$_r$ and \text{GRU}$_e$ denote the GRU networks for the relation classification and entity recognition modules in the interaction layer. Denote $H_r^k$ and $H_e^k$ the task-specific features modeled by the respective \text{GRU}$_r$ and \text{GRU}$_e$ networks at the $k$-th layer. Formally, the outputs $H_r^k$ and $H_e^k$ and shared features $H^k$ at the $k$-th interaction layer is computed as follows:
\begin{equation}
\begin{aligned}
H_{r}^k &= \text{GRU}_r\left(Y_r^{k-1},H^{k-1} | \theta_{\text{GRU}_{r}} \right) \\
H_{e}^k &= \text{GRU}_e\left(Y_e^{k-1},H^{k-1} | \theta_{\text{GRU}_{e}} \right) \\
H^k &= H_{r}^{k} + H_{e}^{k} + H^{k-1} 
\end{aligned}
\end{equation}
where $\theta_{\text{GRU}_{r}}$ and $\theta_{\text{GRU}_{e}}$ are parameters for the \text{GRU}$_r$ and \text{GRU}$_e$ networks respectively. To take advantage of the previous learned explicit  interactions in this network, we allow the network to have a minimum of two layers, i.e, $k=2,3\ldots K$. Hence, for the ER task and RC tasks, the outputs at the $k$-th layer is formulated as:
\begin{equation}
\begin{aligned}
Y_r^k &= C_r(H_r^k) \\
Y_e^k &= C_e(H_e^k)     
\end{aligned}
\end{equation}
\subsubsection{The GRU network}
In the RIN model, we proposed the \text{GRU}$_r$ and \text{GRU}$_e$ networks for the relation classification and entity recognition modules. Formally, for a single word $w$, the \text{GRU}$_e$ network takes the output $y\in Y_e$ and the shared word representation $h\in H$ as inputs and computes the ER task feature vector $h_e\in H_e$. Formally, this can be formulated as:
\begin{equation}
\begin{aligned}
z&=\sigma \left( W_{\rm z}(h\,  \oplus \,y) \right)\\
u&=\sigma \left( W_{\rm u}(h\,  \oplus \,y) \right)\\
\Check{h}&={\rm tanh} \left( W_{\rm o}((u*h) \,\oplus\, y) \right)\\
h_e&=(1-z)*h+z*\Check{h}
\end{aligned}
\end{equation}
where $ \oplus$ is a concatenation operator, $W_z,W_u,W_o$ are learnable parameters of the GRU network. \text{GRU}$_r$ follows the same architecture as \text{GRU}$_e$ to compute the RC task feature vector $h_r\in H_r$ for word $w$. However, for a given word $w_i$, it considers $h_i\in H$ and the vector $y_i$, where $y_i$ is modeled from the set of relation predictions for all word pairs containing $w_i$. We can define this set as $Y_r(w_i)=: \{y_{(i,j)}\in Y_r| w_j\in s\}$. 


\begin{equation}
y_i = \text{maxpool} \left(\,Y_r(w_i)\right),
\end{equation}

where the function \text{maxpool}$(\cdot)$ is a maxpool operation along the dimension.

\subsection{Entity Recognition (ER)}

The ER module $C_e$ attempts to recognize all entities in the text based on the features $H_e$. As an entity may consist of multiple words, we formalize the ER task as tagging each word with an entity label, taking values from \emph{(Begin, Inside, End, Single, Out)} using the BIOES tagging scheme~\cite{DBLP:conf/acl/FuLM19}. Specifically, the ER module classifies each word to one of the five label clusters. The probability distribution  $y$ of word $w$ over these five clusters is calculated based on the ER task feature vector $h_e$ as follows:
\begin{equation}
y={\rm softmax}(W_{\rm e}{h_e}+b_{\rm e}),
\end{equation}
where $\theta_{\rm ER}=\{W_{\rm e},b_{\rm e}\}$ are learnable model parameters.

\subsection{Relation Classification (RC)}
\label{sec:RC-module}
The RC module $C_r$ makes an attempt to identify and extract relational facts from the sentence. Following~\cite{DBLP:conf/acl/FuLM19}, we classify all relations between pairs of words in the sentence based on the features $H_r$. Thus, the relation classification task is interpreted as a binary classification problem, where we identify the truth value of a relational triple $\left<w_i,t,w_j\right>$ by classifying the word pair $(w_i,w_j)$. The task can be regarded as learning the probability distribution $y_{(i,j)}$ for each word pair $(w_i,w_j)$. The value $y_{(i,j)}$ is a probability distribution over the relation types $t\in T$. Thus, $y_{(i,j)}$ is a vector with size $l$, where each dimension is a probability $y_{(i,j)}^t$ of the relational triple  $\left<w_i,t,w_j\right>$ to be factual. We compute $y_{(i,j)}$ for each word pair $(w_i,w_j)$ by performing the following steps:
\begin{equation}
\begin{aligned}
m&=\phi \left( W_{\rm m}(h_i \, \oplus \, h_j) \right) \\
y_{(i,j)}&=\sigma \left( W_{\rm r}m+b_{r} \right)\label{eq:p}
\end{aligned}
\end{equation}
where $h_i,h_j\in H_r$ are the RC task feature vectors for $w_i,w_j\in s$, $ \oplus$ is a concatenation operation, $\phi (\cdot)$ is the ReLU activation function, $\sigma (\cdot)$ is the sigmoid activation function.
$\theta_{\rm RC}=\{ W_{\rm m},W_{\rm r}, b_r\}$ are learnable model parameters. Instead of using a softmax function for classification, as used in \cite{DBLP:conf/acl/FuLM19}, we find that the sigmoid function offers a natural way of identifying multiple relations that may exist between word pairs, solving the overlapping problem more efficiently.

\subsection{Training Objective}
The RIN model ultimately outputs task-specific representations, which are fed into their corresponding ER module and the RC module for predictions. As such, the training objective of RIN is comprised of two parts: the loss function for RC $L_{\rm r}$ and the loss function for ER $L_{\rm e}$. The losses $L_{\rm e}$ and $L_{\rm r}$ are defined as
\begin{equation}
\begin{aligned}
L_{\rm e}(w)&={\rm CrossEntropy}\left({\bar y},  y\right) \\
L_{\rm r}(\left<w_i,t,w_j\right>)&={\rm CrossEntropy}\left({\bar{y}_{(i,j)}^t},{y_{(i,j)}^t}\right)
\end{aligned}
\end{equation}

where ${\bar y}$ and ${\bar{y}_{(i,j)}^t}$ are the respective ground truth values of word $w$ and relational triple $\left<w_i,t,w_j\right>$, and $y$ and $y_{(i,j)}^t$ are the predictions from the ER module ($C_e$) and the RC module ($C_r$) at the $K$-th layer (i.e. the last layer) of RIN. 



The total loss $L$ over all words and relational triples for all sentences is then calculated as follows.
\begin{equation}
L\!=\!\sum_s \!\left(\!\sum_{w\in s}\! L_{\rm e}\!(w)+\!\!\!\!\!\!\!\!\!\!\sum_{w_i,w_j\in s,t\in T}\!\!\!\!\!\!\!\!\!\! L_{\rm r}(\!\left<w_i,t,w_j\right>\!)\!\right)\!
\end{equation}
With gradient based algorithm, we seek to minimize the total loss $L$ over all model parameters $\Theta\!=\!\{ \! \theta_{\text{GRU}_{r}},\theta_{\text{GRU}_{e}},\theta_{\rm RC},\theta_{\rm ER},\theta_{\rm H}$\} ($\theta_{\rm H}$ is the parameters for the BiLSTM network) to achieve good performance for both the ER and RC tasks.  

\begin{table*}[htbp]
\centering
\small
\begin{tabular}{l|l|ccc|ccc}\hline
&&\multicolumn{3}{c|}{NYT} &
\multicolumn{3}{c}{WebNLG}  
\\   \cline{3-8}
Evaluation &{Model} & Prec & Rec & F1 & Prec & Rec & F1\\ \hline 
&OneDecoder  &59.4&53.1&56.0&32.2&28.9&30.5\\ 
&MultiDecoder  &61.0 &56.6 &58.7 &37.7 &36.4 &37.1 \\
Partial Match&OrderRL  &77.9 &67.2 &72.1 &63.3 &59.9 &61.6\\
\cline{2-8}
&RIN$_{\text{w/o interaction}}$&83.9$\pm$0.6&83.1$\pm$0.6&83.5$\pm$0.2&84.9$\pm$0.6&86.3$\pm$0.8&85.6$\pm$0.3\\
&RIN&\textbf{87.2$\pm$0.2}&\textbf{87.3$\pm$0.3}&\textbf{87.3$\pm$0.1}&\textbf{87.6$\pm$0.1}&\textbf{87.0$\pm$0.9}&\textbf{87.3$\pm$0.4}\\
\hline
&NovelTagging  &62.4 &31.7 &42.0 &52.5 &19.3 &28.3 \\
&GraphRel$_{1p}$ &62.9 &57.3 &60.0 &42.3 &39.2 &40.7 \\
&GraphRel$_{2p}$  &63.9 &60.0 &61.9 &44.7 &41.1 &42.9 \\ 
Exact Match&CopyMLT-One   &72.7 &69.2 &70.9 &57.8 &60.1 &58.9 \\
&CopyMLT-Mul   &75.7 &68.7 &72.0 &58.0 &54.9 &56.4 \\
\cline{2-8}
&RIN$_{\text{w/o interaction}}$&77.4$\pm$1.1&76.4$\pm$0.7&76.9$\pm$0.3&75.0$\pm$1.1&73.3$\pm$0.7&74.2$\pm$0.3\\
&RIN&\textbf{83.9$\pm$0.5}&\textbf{85.5$\pm$0.5}&\textbf{84.7$\pm$0.4}&\textbf{77.3$\pm$0.7}&\textbf{76.8$\pm$1.0}&\textbf{77.0$\pm$0.2}\\
\hline
\end{tabular}
\caption{Precision, Recall and F1 performance of different models on the datasets. Results for the compared models are retrieved from their original papers. We report the mean results over five runs and the standard deviation. The best performance is bold-typed.}
\label{table:Result}
\end{table*}

\section{Experiment}\label{experiment}

We conduct experiments to evaluate RIN on two public datasets NYT
~\cite{DBLP:conf/pkdd/RiedelYM10} and WebNLG
~\cite{DBLP:conf/acl/GardentSNP17}. The NYT dataset consists of $1.18$M sentences with $24$ predefined relation types. The WebNLG dataset was created by Natural Language Generation (NLG) tasks, and adapted by~\cite{DBLP:conf/acl/LiuZZHZ18} for a relational triple extraction task. We directly use the preprocessed datasets released by~\cite{DBLP:conf/acl/LiuZZHZ18}~\footnote{https://github.com/xiangrongzeng/copy\_re}. It is worth mentioning that only the tail word of an entity is marked in the preprocessed dataset released by \cite{DBLP:conf/acl/LiuZZHZ18}. To properly distinguish entities, we take a further step of tagging entities with the conventional BIOES tagging scheme as the one used in \cite{DBLP:conf/acl/FuLM19}. We report Precision (Prec), Recall (Rec) and micro-F1 (F1) scores on our model and other recent models~\cite{DBLP:conf/acl/LiuZZHZ18,zeng-etal-2019-learning,DBLP:conf/acl/ZhengWBHZX17,DBLP:conf/acl/FuLM19,zeng2019copymtl} for the {\it Partial Match} task and the {\it Exact Match} task. For our proposed method, we report the mean results over five runs using different random seeds, along with its standard deviation to show the stability of our results. The statistics of datasets are summarized in Table~\ref{tab:data_stat}. Additional experiments on older datasets NYT10~\cite{DBLP:conf/pkdd/RiedelYM10} and NYT11~\cite{DBLP:conf/acl/HoffmannZLZW11} are also performed, and the results are available in the supplementary file. Our results on these datasets show satisfactory performance, generally outperforming previous models on the NYT10 and NYT11. 



\subsection{Partial Match and Exact Match}
Both NYT and WebLG datasets support evaluation for the {\it Partial Match} task and the {\it Exact Match} task.  The Partial Match task only requires the relation and the heads of both subject and object entities of the extracted relational triple  to be correct. For the Exact Match as recently adopted by \cite{DBLP:conf/acl/ZhengWBHZX17,DBLP:conf/acl/FuLM19,zeng2019copymtl}, the extracted relational triple  is considered to be correct if the relation and the heads and tails of the subject and object entities are all correct. Thus, the extracted relational triple completely matches the gold relational triple. 

\begin{table}[h]
\small
    \centering
\begin{tabular}{c|c|c|c}\hline
Dataset& Train& Dev  & Test \\   \hline
NYT  & 56195    & 5000     & 5000     \\ 
WebNLG  & 5019   & 500   & 703  \\
 \hline
\end{tabular}
    \caption{Statistics of NYT and WebNLG}
    \label{tab:data_stat}
\end{table}

\subsection{Implementation Details}
For a fair comparison with previous recent works~\cite{DBLP:conf/acl/LiuZZHZ18}, we use the $100$-dimensional Glove embedding~\cite{pennington2014glove} to represent the word embeddings~\footnote{https://nlp.stanford.edu/projects/glove/}. Part-of-speech (POS) tags are assigned to words using Stanford POS tagger~\footnote{https://stanfordnlp.github.io/CoreNLP/}. We map each POS tag to a randomly initialized 10-dimensional POS embedding.  We concatenate both word and POS embeddings as the input embeddings. For any given sentence, the input embeddings for words are fed to a BiLSTM network to learn a 100-dimensional embedding for each word. We improve learning by using dropout regularization in the input embeddings. The BiLSTM embeddings represent the shared features $H$ in the RIN model. Our model is trained using an Adam optimizer~\cite{kingma2014adam}. The hyper-parameters are set empirically and manually tuned on the development set to select the best model. We implement our model using PyTorch on a Linux machine with a GPU device NVIDIA V100 NVLINK 32GB. Table~\ref{tab:parameter_settting} lists the hyper-parameters of RIN on the datasets. For the relation classification task, we threshold the probabilities of the prediction and return only the relations with probability values $\geq 0.5$. The code for our model will be made available upon acceptance.


\begin{table}[htbp]
    \centering
    \small
    \begin{tabular}{c|c|c|c}
    \hline
         &Hyper-parameter&NYT&WebNLG \\
    \hline
        &$K$&$4$&2\\
        &$d$&$0.1$&$0.1$\\
        Partial Match&$\eta$&$1e^{-3}$&$5e^{-4}$\\
        &bs&$50$&$50$\\
        & epochs &$100$&$150$\\
    \hline
        
        &$K$&$7$&3\\
        &$d$&$0.1$&$0.1$\\
        Exact Match&$\eta$&$1e^{-3}$&$5e^{-4}$\\
        &bs&$50$&$50$\\
        & epochs &$100$&$150$\\
    \hline
    \end{tabular}
    \caption{Hyper-parameter settings of RIN on the datasets ($K$: number of interaction layers, $d$: dropout rate for input embeddings, $\eta$: learning rate, bs: batch size.)}
    \label{tab:parameter_settting}
\end{table}

\subsection{Performance Comparison}
We compare our model with several recent models based on the Partial Match and the Exact Match evaluation tasks.  We also include a baseline model RIN$_{\text{w/o interaction}}$ which excludes the interaction network used in RIN. In RIN$_{\text{w/o interaction}}$, the shared features $H$ modeled by BiLSTM network is directly passed into $C_{e}$ and $C_{r}$ for task-specific predictions. We also compare with several recent models, including the NovelTagging~\cite{DBLP:conf/acl/ZhengWBHZX17}, 
sequence-to-sequence (seq2seq) models such as OneDecoder~\cite{DBLP:conf/acl/LiuZZHZ18}, MultiDecoder~\cite{DBLP:conf/acl/LiuZZHZ18}, and OrderRL ~\cite{zeng-etal-2019-learning}, and MTL-based methods CopyMLT~\cite{zeng2019copymtl}, and GraphRel~\cite{DBLP:conf/acl/FuLM19}.

\begin{figure}[h!]
	\centering
	\renewcommand\tabcolsep{0pt}
	\begin{tabular}{c c}
	\scalebox{0.25}{\includegraphics[]{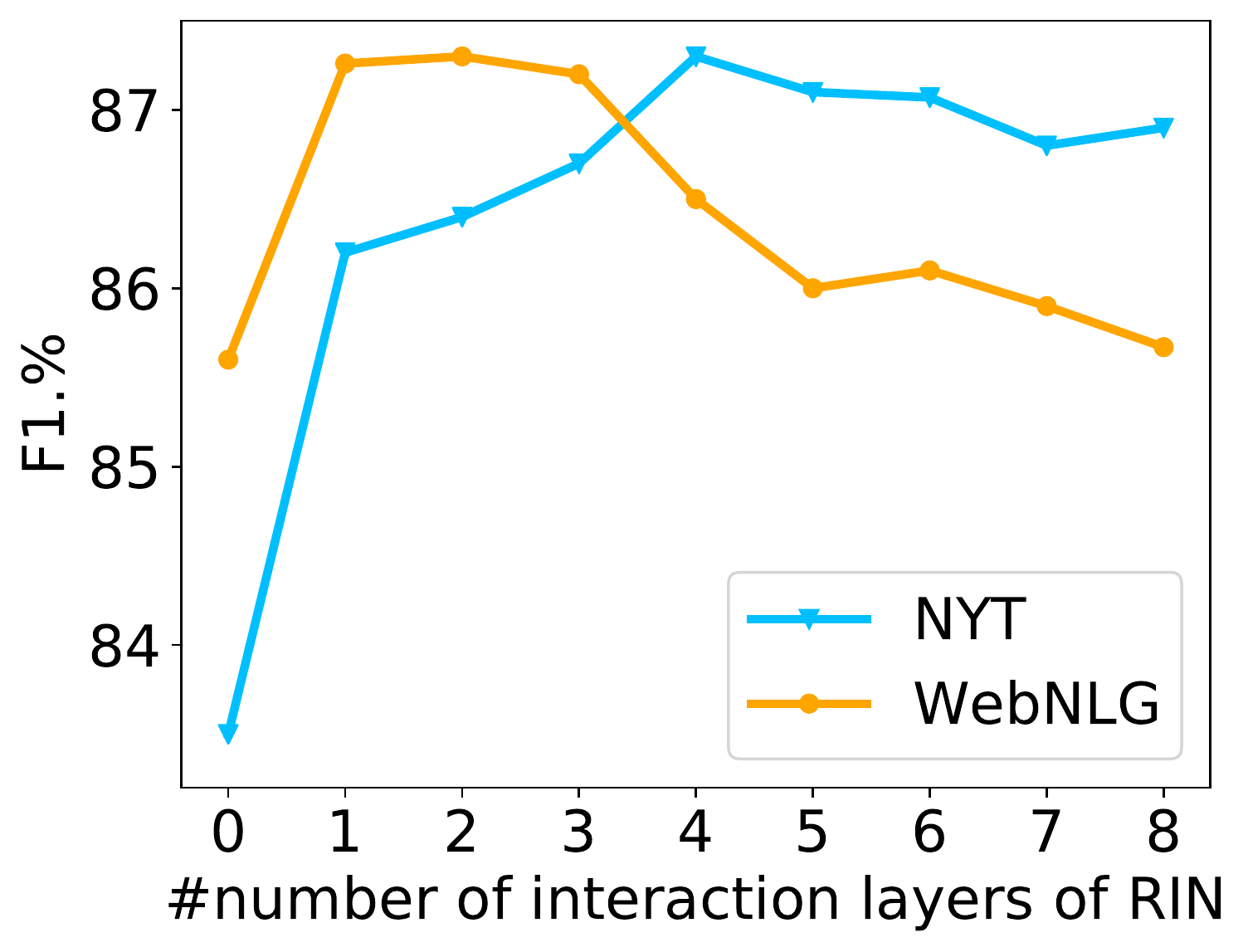}}
	&\scalebox{0.25}{\includegraphics[]{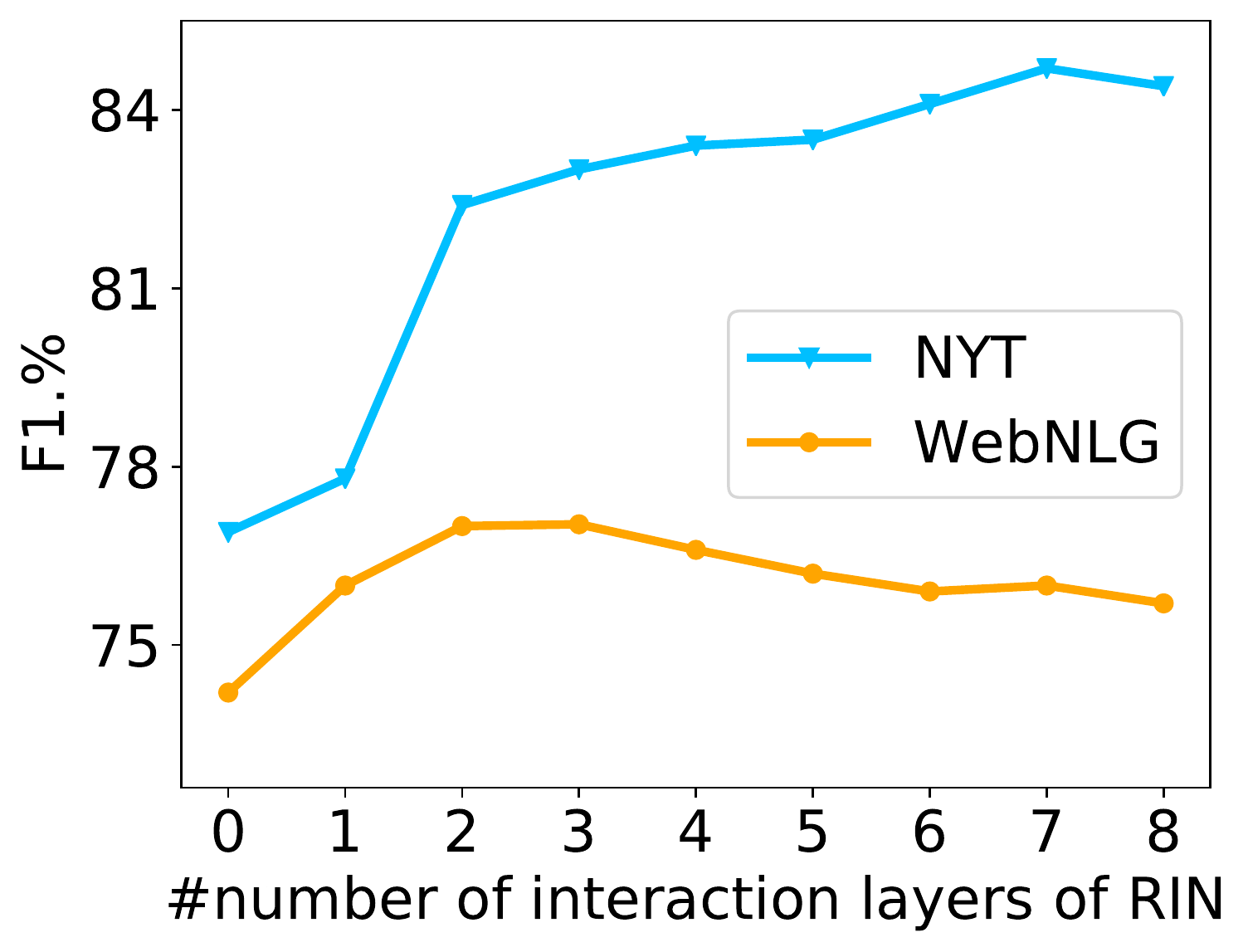}}\\
	(A) Partial Match& (B) Exact Match
	\end{tabular}
	\caption{Curves of F1 performance on different number of interaction layers $K$.}
	\label{fig:updating_rounds}
\end{figure}

\noindent\underline{\bf Partial Match}
Table~\ref{table:Result} shows the performance of different models on the datasets. For the Partial Match evaluation task, it can first be noted that the small standard deviation for our model RIN and its ablated model RIN$_{\text{w/o interaction}}$ shows that our results are stable to an extent on the datasets. Even with the simple structure of RIN$_{\text{w/o interaction}}$, its results outperform the compared methods. In extracting relational facts, our model treats the Partial Match task as a relation classification problem. Whereas the compared methods take a seq2seq based approach to directly extract relational facts in the sentence. The results suggest that our approach may be more effective in identifying the relational facts for this task. It is more interesting to see the performance achieved by RIN. First of all, it can be noted that the model shows a level of stability due to its small standard deviation. Moreover, RIN shows a significant performance boost to the RIN$_{\text{w/o interaction}}$ model, suggesting the importance of dynamically learning the explicit interactions between the ER task and the RC task.


\noindent\underline{\bf Exact Match}
For the Exact Match task, we do not consider the methods~\cite{DBLP:conf/acl/LiuZZHZ18,zeng-etal-2019-learning}, since these methods consider a seq2seq approach in extracting relational triples. Seq2seq methods are able to only decode a single word for an entity. Hence, they will inevitably fail to identify entities with multiple words.

In Table~\ref{table:Result}, we find that our ablated model RIN$_{\text{w/o interaction}}$ consistently outperforms previous models on the two datasets. In a more detailed analysis, we can note that the variants of the GraphRel model~\cite{DBLP:conf/acl/FuLM19} consider the Exact Match task as a relation classification problem which classifies all word pairs in the sentence. In its relation classification module, it exploits a softmax function for the final classification. Hence, the model is not able to address cases where multiple relations exist between a pair of entities. We believe this explains why its results underperforms when compared to RIN$_{\text{w/o interaction}}$. Although CopyMLT and its variants~\cite{zeng2019copymtl} consider a seq2seq based approach to directly extract relational triples, its ER model can identify entities with multiple words and hence can address the Exact Match task. Nonetheless, it fails to outperform our model due to the fact that it uses a seq2seq based approach which we believe to be a more complex method for identifying relational triples. Besides, our main model RIN significantly outperforms RIN$_{\text{w/o interaction}}$ on the two datasets, further proving the importance of the explicit interactions learned between the ER and RC tasks. 


\subsection{Impact of $K$ on the results}
The hyper-parameter $K$ is the number of interaction layers of the RIN model. Thus, $K$ controls the number of times the RIN model attempts to learn explicit interactions between the ER and RC task.  We conduct experiments to study the impact of $K$ on the performance of RIN. We expect that the performance of the model increases as we learn more explicit interactions between the ER task and the RC task. 
\begin{table*}[h!]
    \centering
    \scalebox{0.8}{
    \begin{tabular}{|p{19em}|p{20em}|}
    \hline
     \multirow{6}{17em}{\it{Case1:  A cult of victimology arose and was happily}  \it {exploited by  clever radicals among \textcolor{blue}{Europe}’s Muslims,}  \it {especially certain religious leaders like Imam Ahmad}  \it {Abu Laban in \textcolor{blue}{Denmark} and Mullah Krekar in \textcolor{blue}{Norway}.}}
     & \textbf{Golden}:\it{\textcolor{blue}{Europe, Denmark, Norway}}\\
      & \textcolor{orange}{\it{(Europe, /location/location/contains, Denmark)}} \\
      & \textcolor{orange}{\it{(Europe, /location/location/contains, Norway)}} \\
      & \textbf{RIN$_{\text{w/o interaction}}$}: \it {\textcolor{blue}{Europe, Denmark, Norway}} \\ 
      & \textcolor{orange}{\it{(Europe, /location/location/contains, Denmark)}} \\
      & \textbf{RIN}: \it {\textcolor{blue}{Europe, Denmark, Norway}}\\
      & \textcolor{orange}{\it{(Europe, /location/location/contains, Denmark)}}\\
      & \textcolor{orange}{\it{\textbf{(Europe, /location/location/contains, Norway)}}}\\
      \hline
       \multirow{6}{17em}{\it{Case2: \textcolor{blue}{Scott} (No rating , 75 minutes) Engulfed by nightmares, blackouts} \it{and the anxieties of the age, a Texas woman flees homeland insecurity for a \textcolor{blue}{New York}} \it{vision quest in this acute, resourceful and bracingly ambitious debut film.}}
       &\textbf{Golden}: \it {\textcolor{blue}{Scott, New York}} \\ 
       &\textcolor{orange}{ \it {(York, /location/location/contains, Scott)}}\\
       &\textbf{RIN$_{\text{w/o interaction}}$}: \it{\textcolor{red}{\textbf{Texas}}}, \it {\textcolor{blue}{New York}}\\
       &\textcolor{orange}{ \it {(York, /location/location/contains, Scott)}}\\
       &\textbf{RIN}: \it {\textcolor{blue}{Scott, New York}} \\        &\textcolor{orange}{\it {(York, /location/location/contains, Scott)}}\\
       \hline
    \end{tabular}
    }
    \caption{Case study for RIN and RIN$_{\text{w/o interaction}}$. Entities and relational triples are in blue and orange texts respectively. We mark a wrong prediction with a red text.}
    \label{tab:cases}
\end{table*}
Figure~\ref{fig:updating_rounds} shows the F1 curves of RIN on the datasets for increasing values of $K$. Here, at $K=0$ the RIN model is reduced to RIN$_{\text{w/o interaction}}$. 

We observe that as $K$ increases the performance of RIN increases to an extent up to a point where it overfits. Taking a closer look at the performance on the Partial Match task, we find that RIN$_{\text{w/o interaction}}$ poorly models the interaction between the ER and RC task. By learning explicit interactions using the RIN model, we observe a sharp rise in performance at $K=1$. On the Exact Match task we observe an interesting behaviour of RIN on the NYT and WebNLG dataset. Note that the $60\%$ of entities on the WebNLG are multi-tokens, while $30\%$ of the entities in the NYT dataset are multi-tokens. This means that the Exact Match task is more difficult on the WebNLG dataset, compared to the NYT dataset. As a consequence, RIN finds it difficult to learn explicit interactions on the WebNLG, while it learns much more easily on the NYT as $K$ rises. We observe a sharp rise in performance from the first layer to the second layer on the NYT dataset. The second layer of RIN takes advantage of the original shared features $H$ and the task-specific features of the first interaction layer. Thus, effective learning of the interaction between the two tasks takes place from the second layer. This explains the sharp rise in performance on the NYT dataset for the Exact Match task. 

The results suggests that, to an extent, the proposed RIN model can dynamically filter additional interaction information between the two tasks along the multiple interaction layers.

\subsection{Ablation Experiment}
To examine the contributions of our main model components, we conduct ablation experiments on the NYT and WebNLG datasets. We use the default hyper-parameter settings for the ablated models (see Table~\ref{tab:parameter_settting}). The ablated models are: (1) RIN$_{\text{w/o ER}}$: A RIN model which excludes the task-specific features $H_e$ in the update of the shared features $H$, restricting the RC module from learning from the ER module. (2) RIN$_{\text{w/o RC}}$: A RIN model which excludes the task-specific features $H_r$ in the update of the shared features $H$, restricting the ER module from learning from the RC module. (3) RIN$_{\text{w/o POS}}$: A RIN which only uses the Glove word embeddings as the input embeddings. We also include the ablated model RIN$_{\text{w/o interaction}}$. Table~\ref{tab:ablation} shows the results for the experiment.

\begin{table}[h!]
    \centering
    \small
    \begin{tabular}{l|c|c}
    \hline
    {Model} & {NYT} & {WebNLG} \\
    \hline
    RIN &84.7  &77.0 \\
    \hline
    RIN$_{\text{w/o ER}}$ &83.9 &76.4\\
    RIN$_{\text{w/o RC}}$ &77.3 &76.0 \\
    RIN$_{\text{w/o interaction}}$ &76.9 &74.2
    \\
    RIN$_{\text{w/o POS}}$&84.1 &76.6\\
    \hline
    \end{tabular}
    \caption{F1 performance of different ablation models on the datasets. The Exact Match evaluation is used.}
    \label{tab:ablation}
\end{table}

We find that the performance of RIN deteriorates as we remove critical components. Among the ablated models designed, RIN$_{\text{w/o interaction}}$ performs very poorly on the two datasets, suggesting the importance of learning explicit interactions dynamically between the ER and RC tasks. We also find that RIN$_{\text{w/o ER}}$ marginally underperforms the RIN model, and also showing a better performance when compared to RIN$_{\text{w/o RC}}$. The results suggest that the performance of RIN is heavily dependent on the ER module exploiting information from the RC module. Lastly, the results for RIN$_{\text{w/o POS}}$ suggest that the POS tags does not significantly boost the performance of RIN.



\subsection{Case Study}
We present two case examples from NYT dataset as illustrations to observe the behaviour of the RIN and RIN$_{\text{w/o interaction}}$ models. Table~\ref{tab:cases} shows the results of the study. In the first case example, both RIN and  RIN$_{\text{w/o interaction}}$ correctly extracts all the gold entities in the sentence. But, RIN$_{\text{w/o interaction}}$ captures only the gold relational triple  \emph{\tt (Europe, /location/location/contains, Denmark)}, and misses the gold triple  \emph{\tt (Europe, /location/location/contains, Norway)}.  Given the fact that \emph{\tt (Europe, /location/location/contains, Norway)} overlaps a relational fat, it is important to dynamically learn to capture the complex interaction between the ER and RC tasks. The RIN model takes advantage of its interaction network to identify both gold triples.


In the second case, we observe that both RIN and RIN$_{\text{w/o interaction}}$ correctly extract the relational triple \emph{\tt (York, /location/location/contains, Scott)}. However, RIN$_{\text{w/o interaction}}$ identifies \emph{\tt Texas} as an entity by error while RIN correctly extracts the entity \emph{\tt Scott} and \emph{\tt New York}. The results suggest that RIN is able to leverage information from the RC module to correctly identify entities in the ER module. It is worth noting that we can easily complete the entity \emph{\tt York} in the extracted relational triple by aligning it to the extracted entity \emph{\tt New York}.  


\section{Conclusion}\label{conclusion}

In this paper, we tackle the weakness of existing MTL-based methods proposed for the joint extraction of entities and relation in unstructured text. Specifically, these methods assume that a shared network is sufficient to capture the correlations between the entity recognition task and the relation classification task, and that the shared features derived from this network can be passed into models for the task-specific tasks to make predictions independently. Instead, we show that dynamically learning the interactions between the tasks may capture complex correlations which improves the task-specific feature for classification. We proposed multi-task learning model which allows explicit interactions to be dynamically learned among the sub-tasks. Our experiments on benchmark datasets validates clear advantage over the existing proposed methods. We note that our model can be adapted for other NLP tasks, including aspect level sentiment classification and slot filling. As future work, we intend to explore its application in those fields.

\section*{Acknowledgment}
This work is supported partly by China 973 program (No. 2015CB358700), by the National Natural Science Foundation of China (No. 61772059, 61421003), by the Beijing Advanced Innovation Center for Big Data and Brain Computing (BDBC), by State Key Laboratory of Software Development Environment (No. SKLSDE-2018ZX-17) and by the Fundamental Research Funds for the Central Universities.

\bibliography{emnlp2020}
\bibliographystyle{acl_natbib}

\end{document}


\maketitle

\section{Computing Infrastructure}
Our experiments are executed on a NVIDIA V100 NVLINK 32GB over the Ubuntu 18.04 LTS. We implement our code on Pytorch 1.1.0 library with Python 3.5.4. Our code has been uploaded in the EMNLP2020 submission system and will also be available after acceptance.

\section{Implementation Details}
We use grid search to choose the following four hyperparameter values: learning rate, batch size, dropout rate and number of interaction layers. We search learning rate in [$1e^{-4}$, $5e^{-4}$, $1e^{-3}$], batch size in [$30$, $50$], dropout rate in [$0.5$, $0.6$, $0.7$, $0.8$], and number of interaction layers in [$1$, $2$, $3$, $4$, $5$, $6$]. The final set-ups of these parameters is shown in table~\ref{tab:10_11_parameter_settting}.
\begin{table}[htbp]
    \centering
    \small
    \begin{tabular}{c|c|c|c}
    \hline
         &Hyper-parameter&NYT10&NYT11 \\
    \hline
        &$K$&5&2\\
        &$d$&$0.7$&$0.8$\\
        Partial Match&$\eta$&$1e^{-3}$&$1e^{-3}$\\
        &bs&$30$&$30$\\
        & epochs &$50$&$50$\\
    \hline
        
        &$K$&5&3\\
        &$d$&$0.7$&$0.8$\\
        Exact Match&$\eta$&$1e^{-3}$&$1e^{-3}$\\
        &bs&$30$&$30$\\
        & epochs &$50$&$50$\\
    \hline
    \end{tabular}
    \caption{Hyper-parameter settings of RIN on the datasets ($K$: number of interaction layers, $d$: dropout rate for input embeddings, $\eta$: learning rate, bs: batch size.)}
    \label{tab:10_11_parameter_settting}
\end{table}

\section{Supplemental Experiments}
Proposed works extracting entities and relations jointly perform evaluations on NYT and WebNLG, or NYT10 and NYT11. We chose to perform evaluations on NYT and WebNLG datasets because it was recently released, and  recent models perform evaluations on these datasets. However, we also evaluated our model on the NYT10 and NYT11 to confirm the effectiveness of our approach. Table~\ref{tab:data_10_and_11} shows the dataset statistics.
 \begin{table}[h]
\small
    \centering
\begin{tabular}{c|c|c}\hline
Dataset& Train  & Test \\   \hline
NYT10  & 70339        & 4006     \\ 
NYT11  & 62648    & 369  \\
 \hline
\end{tabular}
    \caption{Statistics of NYT10 and NYT11}
    \label{tab:data_10_and_11}
\end{table}

For a fair comparison, we use the 300-dimensional Glove embedding~\cite{pennington2014glove} which is also used by~\cite{DBLP:conf/aaai/TakanobuZLH19}. We pass the input embeddings into a BiLSTM to learn a 200-dimensional shared representations. Table~\ref{tab:10_11_parameter_settting} lists the main hyper-parameters. We directly use the preprocessed datasets released by~\cite{DBLP:conf/aaai/TakanobuZLH19}~\footnote{https://github.com/truthless11/HRL-RE}. We randomly select 10\% of samples from the datasets as the development set. We compare with several recent models. For the partial match evaluation, we compare with MultiR~\cite{DBLP:conf/acl/HoffmannZLZW11}, FCM~\cite{DBLP:conf/emnlp/GormleyYD15}, SPTree~\cite{DBLP:conf/acl/MiwaB16}, CoType~\cite{DBLP:conf/www/RenWHQVJAH17}, NovelTagging~\cite{DBLP:conf/acl/ZhengWBHZX17}, MultiDecoder~\cite{DBLP:conf/acl/LiuZZHZ18} and HRL~\cite{DBLP:conf/aaai/TakanobuZLH19}. For the exact match evaluation, we compared with~\cite{DBLP:conf/emnlp/LiuRZZGJH17}, LSTM-CRF~\cite{DBLP:journals/ijon/ZhengHLBXHX17} and PA-LSTM-CRF~\cite{DBLP:conf/aaai/DaiXLDSW19}. 
\begin{table*}[h!]
\centering
\small
\begin{tabular}{l|l|ccc|ccc}\hline
&&\multicolumn{3}{c|}{NYT10} &
\multicolumn{3}{c}{NYT11}  
\\   \cline{3-8}
Evaluation &{Model} & Prec & Rec & F1 & Prec & Rec & F1\\ \hline
&MultiR &-&-&-&32.8&30.6&31.7\\
&FCM&-&-&-&43.2&29.4&35.0\\
&SPTree  &49.2&55.7&52.2&52.2&54.1&53.1\\ 
&CoType  &- &- &- &48.6 &38.6 &43.0 \\
Partial Match&NovelTagging  &59.3 &38.1 &46.4 &46.9 &48.9 &47.9\\
&MultiDecoder&56.9&45.2&50.4&34.7&53.4&42.1\\
&HRL&71.4&58.6&64.4&53.8&53.8&53.8\\
\cline{2-8}
&RIN$_{\text{w/o interaction}}$&75.0$\pm$0.8&65.9$\pm$0.4&70.2$\pm$0.2&51.9$\pm$1.4&57.0$\pm$0.5&54.3$\pm$0.8\\
&RIN&\textbf{79.1$\pm$0.7}&\textbf{67.9$\pm$0.6}&\textbf{73.1$\pm$0.1}&\textbf{56.3$\pm$0.8}&\textbf{58.9$\pm$1.4}&\textbf{57.6$\pm$0.2}\\
\hline
&ReHession&-&-&-&41.2&57.3&48.0\\
&LSTM-CRF &-&-&-&\textbf{69.3}&31.0&42.8\\
Exact Match&PA-LSTM-CRF&-&-&-&49.4&\textbf{59.1}&53.8\\
\cline{2-8}
&RIN$_{\text{w/o interaction}}$&72.0$\pm$0.8&59.0$\pm$0.4&64.8$\pm$0.2&50.7$\pm$1.1&55.4$\pm$0.4&53.0$\pm$0.7\\
&RIN&\textbf{77.2$\pm$0.2}&\textbf{65.5$\pm$0.7}&\textbf{70.8$\pm$0.6}&55.3$\pm$0.8&58.5$\pm$0.6&\textbf{56.8$\pm$0.2}\\
\hline
\end{tabular}
\caption{Precision, Recall and F1 performance of different models on the datasets. ``-'' means the result is not reported. For the partial match evaluation task,  the results of the compared models are retrieved from~\cite{DBLP:conf/aaai/TakanobuZLH19}. For the exact match evaluation task, the results of the compared models are retrieved from~\cite{DBLP:conf/aaai/DaiXLDSW19}. We report the mean results over five runs and the standard deviation. The best performance is bold-typed.}
\label{table:10_and_11}
\end{table*}


From Table~\ref{table:10_and_11}, we find that our model shows satisfactory performance on these datasets. Specifically, our model outperforms all models on NYT10 for the Exact Match task and Partial task, including the Partial Match task on the NYT11 dataset. Our model also shows competitive performance with LSTM-CRF~\cite{DBLP:journals/ijon/ZhengHLBXHX17} and PA-LSTM-CRF~\cite{DBLP:conf/aaai/DaiXLDSW19}, outperforming these methods on the F1 score. Although, we do not show this performance in our manuscript, the results goes further to support our idea of dynamically learning the interactions between the entity recognition task and the relation classification task.

\bibliography{emnlp2020}
\bibliographystyle{acl_natbib}